\def\year{2021}\relax
\documentclass[letterpaper]{article} 
\usepackage{aaai21}  
\usepackage{times}  
\usepackage{helvet} 
\usepackage{courier}  
\usepackage[hyphens]{url}  
\usepackage{graphicx} 
\usepackage{natbib}  
\usepackage{caption} 
\usepackage{xcolor}

\newcommand{\argmax}{argmax}
\newcommand{\argmin}{argmin}

\usepackage{amsmath}
\usepackage{amsfonts}
\usepackage{amssymb}

\usepackage{amsthm}

\usepackage{caption}
\usepackage{subcaption}

\theoremstyle{definition}
\newtheorem{definition}{Definition}[section]
\usepackage{eso-pic}

\usepackage{lineno}
\nocopyright
\title{
A Bayesian Account of Measures of Interpretability \\ in Human-AI Interaction
}

\author{
Sarath Sreedharan$^1$ $\cdot$ 
Anagha Kulkarni$^1$ $\cdot$ 
Tathagata Chakraborti$^2$ \\
\Large {David E. Smith $\cdot$ Subbarao Kambhampati$^1$}\\[1ex]
\textnormal{$^1$Arizona State University $\cdot$ $^2$IBM Research AI}
}

\begin{document}

\maketitle

\begin{abstract}
Existing approaches for the design of interpretable agent behavior consider different measures of interpretability in isolation.
In this paper we posit that, in the design and deployment of human-aware agents in the real world, notions of interpretability are just some among
many considerations; and the techniques developed in isolation lack two key properties to be useful when considered together: they need to be able to
1) deal with their mutually competing properties; and 2) an open world where the human is not just there to interpret behavior in one specific form.
To this end, we consider three well-known instances of interpretable behavior studied in existing literature -- namely, explicability, legibility, and 
predictability -- and propose a revised model where all these behaviors can be meaningfully modeled together. 
We will highlight interesting consequences of this unified model and motivate, through results of a user study, why this revision is necessary.

\end{abstract}

\section{Introduction}

A crucial aspect of the design of AI systems that are capable of working alongside humans is the synthesis of interpretable behavior \cite{gunning2019darpa, langley2017explainable}. 
Existing works in this direction \cite{chakraborti2019explicability} 
explore behaviors that instigate a desired change in the human's mental state 
or conform with her current mental state so as to not require explicit 
communication. 
Three distinct notions of interpretability 
can be seen in prior work: 
{\em legibility} (the agent signalling its objectives through behavior);
{\em explicability} (agent behavior that conforms with the human's expectation); and 
{\em predictability} (agent behavior that is easier to anticipate). 
These notions of interpretability can each improve human-AI 
collaborations along different dimensions. 
If you know your robot's objectives (legibility) and can anticipate 
its future behavior (predictability), 
you can plan around it or even exploit it; while in conforming to your 
expectations (explicability), it can avoid surprising you and 
adversely affect the fluency of collaboration. 

Despite much progress in the community in understanding and modeling 
these behaviors individually, 
there has not been any consideration for the effect of one behavior 
on the other in terms of their defining properties. 
Authors in \cite{dragan2015effects} showed how legibility and predictability affects a human observer, engaged in 
a collaborative task, 
though the work looked at the merits of these metrics in isolation, and pitched against one another.
However, in the design and deployment of AI agents that can
collaborate with humans, 
there is no such thing as a ``legible agent'' or an ``explicable agent'' -- 
there are only agents that are human-aware and can, 
among many other considerations such as the modeling of collaborative behavior 
and joint plans with teammates, also exhibit behavior that is interpretable 
to the humans in the loop.
We showed previously how, in the context of human-robot 
teaming \cite{zhang2015human} and in the context of an 
embodied agent in an 
instrumented workspace \cite{ai-comm},
human-aware behaviors designed in isolation -- in those 
cases, proactive support -- may not always bear out in the 
context of a general interaction that is not specifically
geared for that behavior to flourish.

If we desire to create agents that can reason about all of these measures simultaneously, and freely choose to be legible, predictable or explicable, then we will require a way to meaningfully measure all the various measures together.
But considering these differing notions of interpretability at the same time has two immediate consequences -- 
1) The behaviors are defined under competing assumptions -- 
e.g. legible behavior is defined for cases where the observer has multiple hypotheses about the agent which includes the true robot model, while predictable and explicable measures assume observers with a single hypothesis about the agent. For predictability, they assume this hypothesis matches the robot model, while explicability does not make this assumption; and 
2) Their formulations are disparate -- e.g. while predictability and legibility rely on a Bayesian formulation, explicability relies on a formulation based on plan distances.
This means that generalizing these measures to settings where they are all compatible with each other is non-trivial.
As we discussed before, each of these behaviors have their unique 
value propositions but they are, unfortunately, not realizable in unison as per the state of the art.

In this paper, we will show how these measures can in fact be compatible with each 
other, with a revised formulation of interpretability of behavior 
best understood in terms of a Bayesian reasoning process at the human's end where 
they are just trying to understand the robot's model and future plans from 
observed behavior. We will see that a crucial element of this unifying framework
is the presence of the human's belief that they may be wrong about the robot model.
This is quite natural in human-AI interactions, e.g. due to the lack of 
the human's confidence about their knowledge of the agent 
or from their belief that the agent may have malfunctioned. 
As we will show later, the introduction of this belief (e.g. by just 
adding some clutter in the environment) will result in previously legible
behavior becoming both inexplicable and illegible.
We will also show how our framework,
is consistent with previously studied characteristics of interpretability measures, 
but can also correctly predict properties of explicability that have not been previously studied. 
Following is a summary of contributions:

\subsubsection*{Summary of Contributions}


\begin{enumerate}
\item We propose a unified framework where all three interpretability 
measures can be modeled together.
\item We map each measure to a specific phenomena at the observer’s end and propose a single reasoning process -- modeled 
as a Bayesian process -- that captures all 
the effects of agent behavior on the observer’s mental model.
\begin{itemize}
\item[-] 
To this end, we show that the ability to model an ``unknown'' state
is critical to the unification of these competing interpretability metrics.
\item[-] 
We also show how the unification leads to new behaviors -- affected 
by multiple possible behaviors and multiple possible mental models -- that 
existing frameworks cannot model but do bear out in reality. 
\end{itemize}
\item We validate through user studies the above novel properties 
of the proposed framework.
\end{enumerate}

In the discussion section, we also provide a sketch of how this new formulation can be used for planning and further discuss how we can leverage communication to boost these interpretability properties.

\section{Background}
\label{background}

Through most of the discussion we will be agnostic to the specific models used to represent the agent. We will also use the term model in a general sense to include information not only about the actions that the agents are capable of doing and their effects on the world but to also include information on the current state of the world, the reward/cost model and any goal states associated with the problem. 
For certain cases we will assume that the model itself could be parameterized and will use $\theta_i(\mathcal{M})$ to characterize the value of a parameter $\theta_i$ for the model $\mathcal{M}$.

Since we are interested in cases where a human is observing an agent acting in the world, we will mainly focus on agent behavior (instead of plans or policies).
A behavior in this context will consist of a sequence of state, action pairs $\tau$, which we refer to as a {\em trace}.  
The likelihood of the sequence given a model will take the form $P_{\ell}: \mathbf{M}\times \mathcal{T} \rightarrow [0,1]$, where $\mathbf{M}$ is the space of possible models and $\mathcal{T}$ the set of  behavioral traces the agent can generate. 

While we will try to be agnostic to likelihood functions, a fairly common approach \cite{fisac2018probabilistically,baker2007goal}
is a noisy rational model based on the Boltzmann distribution:
$P_{\ell}(M, \tau) \propto e^{-\beta \times C(\tau)}$.



For the human-aware scenario, we are dealing with two different models \cite{dragan2017robot,chakraborti2018foundations,reddy2018you}: the model that is driving the agent behavior (denoted $\mathcal{M}^R$) and the human's belief $\mathcal{M}^R_h$ about it.
We make no assumptions about whether these two models are represented using equivalent representational schemes or use the same likelihood functions. 
This setup assumes that while the human may have expectations about the robot's model, she may have no expectation about its ability 
to model her, thereby avoiding additional nesting. 

\subsection{Existing Interpretability Measures}

We will now provide a brief overview of different 
interpretability measures of interest, following those 
laid out in \citet{chakraborti2019explicability, macnally2018action} (with some generalizations allowed to transfer binary concepts of interpretability
to more general continuous scores).

\subsubsection{Legibility ($\mathcal{L}$)} 
Legibility was originally formalized 
\cite{dragan2013legibility}
as the ability of a behavior to reveal its underlying objective. 
This involves a human who is considering a set of possible goals ($\mathbb{G}$) of the agent and is trying to identify the real goal 
by observing its behavior. 
Legibility is thus formalized as the 
maximization of the probability of the real goal
through behavior:
\begin{align}
\hat{\tau}^{*}_{\mathcal{L}} = \argmax_{\hat{\tau}} P(g^R| \hat{\tau})
\end{align}
where $g^R$ is the real goal of the agent and $\hat{\tau}$ 
is the current prefix. 
While originally introduced in the context of motion planning, 
this was later adapted to task planning by
\citet{macnally2018action}, and 
generalized to implicit communication of beliefs 
when the human has partial observability by \citet{kulkarni2019unified} 
as well as implicit communication of any model parameter
(as opposed to just goals) by \citet{SZ:MZaamas20}.

\subsubsection{Explicability ($\mathcal{E}$)} 
A behavior is explicable if it meets the human's expectation 
from the agent for the given task \cite{zhang2017plan}. 
In the binary form this is usually taken to mean that the behavior is explicable if it is one of the plans the human would have expected the agent to generate \cite{balancing}. 
In the more general continuous form, 
this notion can be translated to be proportional 
to the distance between the observed trace and the 
closest expected behavior \cite{kulkarni2019explicable}: 
\begin{align}
\tau^{*}_{\mathcal{E}} = \argmin_{\tau} \delta(\tau, \tau^{E}_{\mathcal{M}^R_h})
\end{align}
where $\delta$ is some distance function between two plans and $\tau^{E}_{\mathcal{M}^R_h}$ is the closest expected behavior for the model $\mathcal{M}^R$. As specified in \cite{chakraborti2019explicability}, we can extend this to behavior prefixes as following
\begin{align}
\hat{\tau}^{*}_{\mathcal{E}} = \argmin_{\hat{\tau}} \delta(\textsf{Compl}^{\mathcal{M}^R_h}(\hat{\tau}), \tau^{E}_{\mathcal{M}^R_h})
\end{align}
where $\textsf{Compl}^{\mathcal{M}^R_h}(\hat{\tau})$ represents the completion of the prefix $\hat{\tau}$ under the model $\mathcal{M}^R_h$.\footnote{\cite{chakraborti2019explicability} hypothesizes
the possibility of using optimistic and pessimistic completions. 
Our framework will be considering 
an expectation over all possible completions.}
While there is no consensus on the distance function or expected behavior, a reasonable possibility for the expected set may be the set of optimal 
plans \cite{balancing} and the distance can be the cost difference \cite{kulkarni2020designing}.

\subsubsection{Predictability ($\mathcal{P}$):} 
This corresponds to the human's ability to correctly 
predict the completion of an observed behavior prefix \cite{fisac2020generating}. 
Now the goal of the agent is to choose a behavior prefix such that:
\begin{align}
    \hat{\tau}^{*}_{\mathcal{P}} = \argmax_{\hat{\tau}} P(\tau'| \hat{\tau}, \mathcal{M}^R_h)
\end{align}
where $P(\tau'| \hat{\tau}, \mathcal{M}^R_h)$ is the probability of a future behavior $\tau'$ given an observed prefix $\hat{\tau}$ under the model $\mathcal{M}^R_h$.


\subsection{An Illustrative Example}
\label{example}

\begin{figure}[tbp!]
\centering
\includegraphics[width=\columnwidth]{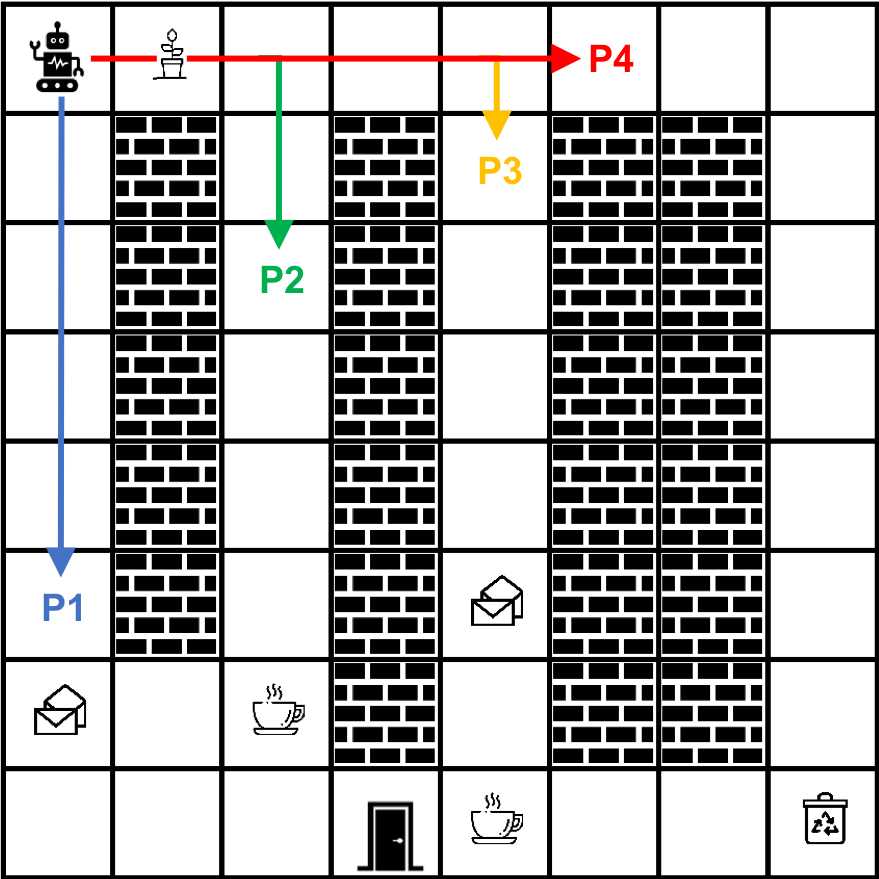}
\caption{
An illustrative example 
of different interpretable behaviors in the 
office robot domain. In this grid, the robot only moves in three directions: down, 
left and right; and it does not revisit a cell.
}
\label{fig:office-1}
\end{figure}

We will consider the operations of a robotic office assistant 
as our running example (Figure \ref{fig:office-1}). 
We will use variations of this domain 
for the user studies as well, 
with changes made mainly to highlight specific properties
of interpretability.

The robot can perform various repetitive tasks in the office, including
picking up and delivering various objects to employees,
emptying trash cans, and so on.
Unlike a standard gridworld scenario, here the robot can only move in three directions: down, 
left and right; as well as it does not revisit a cell.
You, as the floor manager, are tasked with observing the robot 
and making sure it is working properly. 
Given you have seen the robot in action previously, 
you have come to form expectations about the robot's 
capabilities and common tasks it generally pursues, 
though you may not know them for certain: 
e.g. you may think that the goals of the robot are 
either to deliver coffee or to deliver mail to a room
(represented by the door), 
though there may be other possibilities 
that you have not considered. 

Now let us say the robot starts following prefix $P4$. 
At this point having seen the first 5 steps of prefix $P4$, it is not clear what exactly the robot is trying to do -- 
Is it going to empty trash? 
Is it trying to fetch coffee? 
Is it going there for some other reason not currently known to you? 
Here the robot is actually trying to convey that it is going for the coffee.
As per the existing notion of legibility introduced above (which ignores other objects and assumes mail and coffee are the only two possible goals):
prefix $P4$ would have been chosen by the robot
since it makes mail less likely than $P2$ does (in fact, in this domain since the robot can't go up it can no longer reach the mail once it executes $P4$), 
and therefore coffee becomes more legible with $P4$. 
However, given that you did not know that these were the only two
possibilities the robot was considering, the 
notion of legibility as it is defined in prior literature, cannot be used. 

Furthermore, if you did in fact know that those were the only
two possible goals, then the behavior does become legible but 
at the same time would also be classified as inexplicable as per
the existing definition of explicability. 
This highlights how prior literature on explicability, legibility, 
and predictability is deficient when considered together and when 
humans may have multiple hypotheses.
In the rest of the paper, 
we will propose a revised formulation of these concepts so as to 
provide a comprehensive framework for the design of interpretable 
agent behavior.


\section{A Unified Framework}

The measures discussed above in one way or another reason 
about the effect of agent behavior at the observer's end. Thus an 
important part of the proposed reasoning framework would be to 
reason about such effects in a unified form. 

\begin{figure}
\centering
\includegraphics[width=0.9\columnwidth]{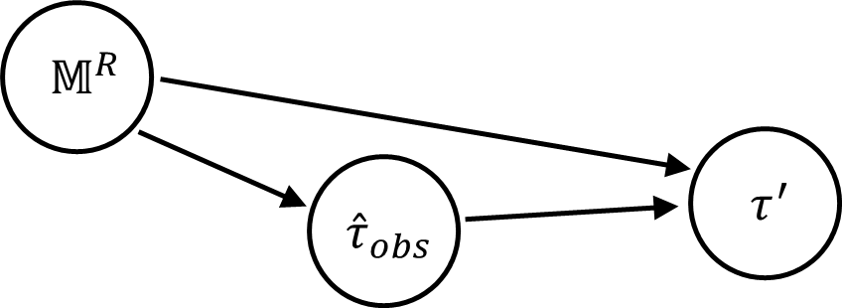}
\caption{Graphical representation of the human's model.}
\label{fig:model}
\end{figure}

In order to do so, we adopt a Bayesian model of the observer's reasoning process (Figure \ref{fig:model}). 
This is motivated by both the popularity of such models in previous works in observer modeling and existing evidence to suggest that people do engage in Bayesian reasoning \cite{l2008bayesian}. 
The node $\mathbb{M}^R$ represents possible models the human thinks 
the agent can have, 
$\hat{\tau}^{obs}$ corresponds to the behavior prefix that they observed, 
and $\tau'$ corresponds to possible completions of the 
prefix.\footnote{
Note that, in this work, we focus on quantifying these properties 
for one shot or episodic interactions only, rather than longitudinal ones.
Later, in Section \ref{conclusion}, we will discuss how
to extend these measures to longer term interactions.
}

In addition to explicit models that the human thinks are possible
for the agent,
we also allow for the possibility that the human may 
not know the agent's model at all.
As we will see later, this is a prerequisite for modeling the notion of explicability as this would correspond to the hypothesis that the human would attribute to unexpected or surprising agent behavior. 
This is also a significant departure 
from existing frameworks, and we will see in the course of this discussion
how this becomes crucial for modeling the different interpretability 
measures together.
With respect to our running example, this applies when 
the human may not be completely familiar with all the details of the robot, or may
allow for the possibility that the robot has malfunctioned when it behaves unexpectedly.
We will incorporate this assumption by adding a special model 
$\mathcal{M}^0$ to the set of models in $\mathbb{M}^R$. 
We represent $\mathcal{M}^0$ using a high entropy model: i.e.
the likelihood function related to this model will assign a small but equal likelihood to any of the possible behaviors. 
This can be viewed as a model belonging to a random agent. 
We will assume that the human in the loop, by default,
assigns smaller priors to $\mathcal{M}^0$ than other models. 


\subsection{Revised Interpretability Measures}
\label{revised}

We assert that the observation of a behavior leads to the human 
updating both their beliefs about the agent model and possible future actions 
of the agent. In the following discussion, we will show how 
we can map each interpretability score (as well as some other related scores relevant to human-aware behavior) to this revised reasoning model. 

\begin{definition}{\bf Explicability:}
\label{defn:explic}
The explicability score of a prefix 
is directly proportional to the probability assigned to all models 
that are not $\mathcal{M}^0$:
\begin{align}
\mathcal{E}(\hat{\tau}_{obs}) \propto \Sigma_{\mathcal{M} \in \mathbb{M}\setminus\{\mathcal{M}^0\}} P(\mathcal{M}|\hat{\tau}_{obs})  
\end{align}
\end{definition}

Intuitively, an explicable behavior indicates that the human
has been able to ascribe the observed behavior to one of the possible
models she thinks the agent has, and hence there is nothing to 
be surprised about. 

While at first sight, the definition might look 
unconnected to the distance based formulation 
discussed in Section \ref{background}, 
these formulations actually turn out to be 
equivalent under certain assumptions. 
Let us consider the set of possible models in $\mathbb{M}^R$ 
to consist of just $\mathcal{M}^0$ and another model $\mathcal{M}^R_h$. 
Then the explicability score will be:
\begin{align}
\mathcal{E}(\hat{\tau}_{obs}) \propto  P(\mathcal{M}^R_h|\hat{\tau}_{obs})\\
\propto  P(\hat{\tau}_{obs}|\mathcal{M}^R_h) * P(\mathcal{M}^R_h)
\end{align}
If $\hat{\tau}_{obs}$ is the entire 
plan\footnote{We considered full observations here to compare with earlier works which have all considered complete plans. In case of prefixes, the likelihood of the prefix can be calculated by marginalizing over the likelihood of all possible completions of the prefix.}, 
then $P(\hat{\tau}_{obs}|\mathcal{M}^R_h)$ is the same as the 
likelihood function described earlier, which gives us:
\begin{align}
\mathcal{E} (\hat{\tau}_{obs}) \propto  P_{\ell}(\mathcal{M}^R_h, \hat{\tau}_{obs}) * P(\mathcal{M}^R_h)
\end{align}
Let us consider two plausible likelihood models. 
First, for a normative model where the agent is expected to be optimal,
$P_{\ell}(\mathcal{M}^R_h, \hat{\tau}_{obs})$ is either $\frac{1}{m}$ ($m$ being the number of optimal plans) leading to high explicability or $0$ for not explicable. This is the original binary explicability formulation used by \citet{balancing, chakraborti2019explicability}. 
Another possible likelihood function is a noisy rational model  \cite{fisac2020generating} given by:
\begin{align}
P_{\ell}(\mathcal{M}^R_h, \hat{\tau}_{obs}) \propto e^{ - \beta \times C(\hat{\tau}_{obs})} \propto e^{ \beta \times C(\tau^{*}) - C(\hat{\tau}_{obs})}
\label{equ:exp}
\end{align}
where $\tau^{*}$ is an optimal behavior, $C(\hat{\tau}_{obs}) \geq C(\tau^{*}) \geq 0$, and $\beta \in \mathbb{R}$  
is a parameter that reflects level of determinism in the users choice of plans \cite{baker2009action}.
This maps the formulation to the distance based definition
as in \citet{kulkarni2020designing} where the distance is defined on the cost. 
We can also recover the earlier normative model by setting $\beta \rightarrow \inf$ and model $\mathcal{M}^0$ by setting $\beta=0$.

Regardless of the specific formulation, explicability is a measure that reflects the user's understanding of the robot behavior generation process (which includes both it's perceived model and its computational component). 
Earlier formulations rely on using the space of expected plans 
as a proxy of this process. 
This is further supported by the fact that, the works that have looked at updating the
human's perceived explicability value of a plan do so by providing information 
about the model and not by directly modifying the human's understanding of the 
expected set of plans \cite{explain}.

An interesting side-effect of a probability based explicability formulation 
is that irrespective of the exact details of the likelihood function, 
the probability of behavior and hence the explicability score is affected by the other plans. For example, consider two scenarios, 
one where $\mathbb{M}^R$ contains $\mathcal{M}_1$ and $\mathcal{M}^0$ and 
another where it contains $\mathcal{M}_2$ and $\mathcal{M}^0$. 
Now consider a behavior trace $\tau$ such that it is equidistant 
from an optimal plan in both models $\mathcal{M}_1$ and $\mathcal{M}_2$. 
Even though they are at the same distance, the trace may be more explicable in the first scenario than in the second, if the second scenario allows for more traces that are closer: i.e. 
in the second scenario these closer plans should have higher probability. 
We argue that this makes intuitive sense for explicability since the user should be more surprised in the second scenario as the agent would have ignored more better behaviors. 
To the best of our knowledge, this is first work to model this 
property of explicability. Thus:

\subsubsection{Property 1} \textit{Explicability of a trace is dependent 
not only on the distance from the expected plans but also on the presence or 
absence of plans close to the expected plans.}

\vspace{5pt}
We will see in Section \ref{hypo1} how this property bears out
in practise, through a user study.

In a more general setting with multiple possible models, 
if we have models with the same prior belief, 
then the formulation makes no difference between plans that work in both models equally well versus ones that works in individual models. 
In essence, the plan remains explicable whether or not the observation leads to all the probability being assigned to a single model versus where they may be distributed across multiple models.
While the exact values would depend on the likelihood function, in the office robot scenario our formulation would assign high explicability scores (need not be the same) to both $P1$ and $P2$. 
For $P1$, the probability mass would be distributed across the two possible hypothesis corresponding to the two goals, while for $P2$ the probability mass is centered around the model corresponding to the goal to fetch coffee. Thus: 

\subsubsection{Property 2} \textit{Explicability is agnostic to whether it is supported by multiple models or by a single one.}

\vspace{5pt}
As with Property 1, we will evaluate how this property bears out 
with humans in the loop in Section \ref{hypo2}.


\begin{definition}{\bf Legibility:}
The legibility score of a prefix for a specific model parameter 
set\footnote{We went with a definition with model parameters instead of 
an explicit set of models, 
since this is currently the most general definition of legibility 
being used in existing literature \cite{SZ:MZaamas20} -- this allows for the possibility that the human might have multiple models in their hypothesis set that may share the same parameters and hence require different approach from just differentiating between models.} 
is directly proportional to the probability of the human's belief in that parameter's value in the true model: 
\begin{align}
\mathcal{L}^{\theta}(\hat{\tau}_{obs}) \propto P(\Theta = \Theta(M^R)|\hat{\tau}_{obs}) \\\propto \Sigma_{\mathcal{M} \in \mathbb{M}\setminus\{\mathcal{M}^0\} ~\textrm{Where}~ \Theta(\mathcal{M}^R)=\Theta(\mathcal{M})} P(\mathcal{M}|\hat{\tau}_{obs})  
\label{equ:leg}
\end{align}
\end{definition}

While this may appear to be a direct generalization of the legibility 
descriptions previously discussed, there are some important points of departure. First, there is no assumption that the actual parameter being 
conveyed or the actual robot model is part of the hypothesis set being 
maintained by the user. Thus it is not always guaranteed that a 
high legibility score can be achieved. Also, note that the parameter is not 
tied to a single model in the set. 

Finally, the presence of $\mathcal{M}^0$ with non-zero prior distribution 
would affect what constitutes legible behavior. Earlier works have 
an explicit assumption that the human is certain that the robot's model is 
one of the few they are considering: i.e. the prior assigned to 
$\mathcal{M}^0$ is zero. This means in many cases existing approaches for 
legible behavior generation can create an extremely circuitous route that is 
more likely in the robot model than others. This means that regardless of how suboptimal the plan is in the robot model (or a true parameter value), given its even lower probability in other models (or for other parameter values) the robot model will get assigned higher posterior probability and thus higher legibility score.
For example in Figure \ref{fig:office-1}, a legible planner might select the prefix $P4$ highlighted in red
in order to reveal the goal of delivering coffee, even though that corresponds to an extremely
sub-optimal plan given the set of possible plans. 
Incorporating $\mathcal{M}^0$ would make sure that 
this path would be given lower score than others. 

Note the similarities between Equation \ref{equ:leg} and Equation \ref{equ:exp}. Both legibility and explicability requires the probability to be distributed across non $\mathcal{M}^0$ models, but legibility requires the human's belief to correspond to true robot model. This is not required in the case of explicability, where the formulation only require the human to have at least one model that could explain the given behavior.
Thus:

\subsubsection{Property 3} \textit{Inexplicable plans are also illegible.}

\vspace{5pt}
We will see in Section \ref{hypo3} to what extent this property
holds in a user study.

\begin{definition}{\bf Predictability:}
Predictability is directly proportional to the 
probability that the completion of an observed prefix in the human's
reasoning process is the same as the completion 
$\tau$ considered by the agent.
\begin{align}
\begin{split}
\mathcal{P} (\tau, \hat{\tau}_{obs}) \propto  P(\tau'=\tau |\hat{\tau}_{obs})\\
\propto  \sum_{\mathcal{M} \in \mathbb{M}} P(\tau'=\tau |\hat{\tau}_{obs}, \mathcal{M}) \times P(\mathcal{M})    
\end{split}
\end{align}
\end{definition}
This is more or less a direct translation of the predictability measure to 
this more general setting -- e.g earlier works \cite{fisac2020generating}
consider a single possible model.
One interesting point to note here is that predictability only optimizes for 
the probability of the current completion, which allows for the system to 
choose unlikely prefixes for cases where the agent is only required to achieve required levels of predictability after a few steps. Going back to the office-robot, the five step prefix $P3$ has high predictability even though 
it is not explicable or legible.

\subsection{Deception and Interpretability}
\label{adversarial}

The interpretability measures being discussed involve leveraging reasoning processes at the human's end to allow them to reach specific conclusions. 
At least for legibility and predictability, the behavior is said to exhibit a 
particular interpretability property only when the conclusion lines up with 
the ground truth at the agent's end. Though as far as the human is considered,
they would not be able to distinguish between cases where the behavior is 
driving them to true conclusions or not. 
This means that the mechanisms used for interpretability could be easily leveraged to perform behaviors that may be adversarial \cite{chakraborti2019explicability}.
Two common classes of such behaviors are deception and obfuscation. 
Deceptive behavior corresponds to behavior meant to convince the user of incorrect information about the agent model or its
future plans \cite{masters2017deceptive}:
\begin{align}
\mathcal{D}^{\mathbb{M}}(\hat{\tau}) \propto  - P(\mathcal{M}^R|\hat{\tau})
\end{align}


Adversarial behaviors meant to confuse the user
are either inexplicable plans that increase the posterior
on $\mathcal{M}^0$ or, plans that 
actively obfuscate \cite{keren2016privacy,kulkarni2019unified}:
\begin{align}
\mathcal{O}^{\mathbb{M}}(\hat{\tau}) \propto H(\mathcal{M}|\hat{\tau})
\end{align}

This is proportional to the conditional entropy of the model distribution given the observed behavior.


With explicability, the question of deceptive behavior becomes 
interesting, since explicable plan generation is really useful
when the agent's model is not part of human's hypothesis set and explicit communication is not possible for the agent. 
With our formulation, the agent can stick to generating optimal plans in those possible models and those would stay explicable. 
This can be construed as deceptive behavior as it is reinforcing incorrect notions about the agent's model. Such plans would have a high deceptive score per the 
formulation above (since $P(\mathcal{M}^R|\hat{\tau}) = 0$). 
One can argue that explicable behaviors are white lies in such scenarios as 
the goal here is just to ease the interaction and the behavior is not driven 
by any malicious intent. In fact, one could even further restrict the 
explicability formulation to a version that only lies by omission
by restricting the agent to generate only optimal behavior in the agent 
model: i.e. among the behaviors that are optimal in its model, 
the agent chooses one that best aligns with human's expectation.

\section{Evaluation}

We now report on user studies performed to validate the new
properties of our formulation, enumerated in Section \ref{revised}. 
The properties already established in the existing literature (which we subsume) are not the focus of our evaluation.



\subsubsection{Hypothesis 1} (c.f. Property 1) \textit{Explicability of a trace will decrease when we allow for additional traces in the model that are closer to expected behaviors.}

\subsubsection{Hypothesis 2} (c.f. Property 2) \textit{A trace that is likely in only a single model versus one that is likely in multiple models will have similar explicability score provided they have similar 
cumulative probability $\Sigma_{\mathcal{M} \in \mathbb{M}\setminus\{\mathcal{M}^0\}} P(\mathcal{M}|\hat{\tau}_{obs})$.}

\subsubsection{Hypothesis 3} (c.f. Property 3) \textit{If a trace has low explicability score per Definition \ref{defn:explic}, then the trace 
will also have low legibility score.}

\vspace{3pt}
We will validate each of these through between-subject and within-subject
studies using a variation of the office scenario 
discussed in Section \ref{example}.
Participants for the studies were recruited from Amazon Mechanical Turk \cite{crowston2012amazon} and for each condition we also had a filter question designed to verify that the participants correctly understood the instructions. All results listed below were calculated on the submissions that had correct answers for the filter questions. 


\begin{figure}[!t]
\centering
\begin{subfigure}[b]{\columnwidth}
\centering
\includegraphics[width=\columnwidth]{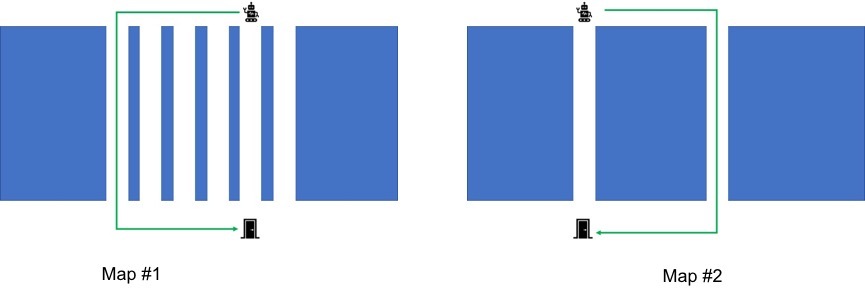}
\caption{Hypothesis 1.}
\label{study1}
\vspace{5pt}
\end{subfigure}
\begin{subfigure}[b]{\columnwidth}
\centering
\centering
\includegraphics[width=\columnwidth]{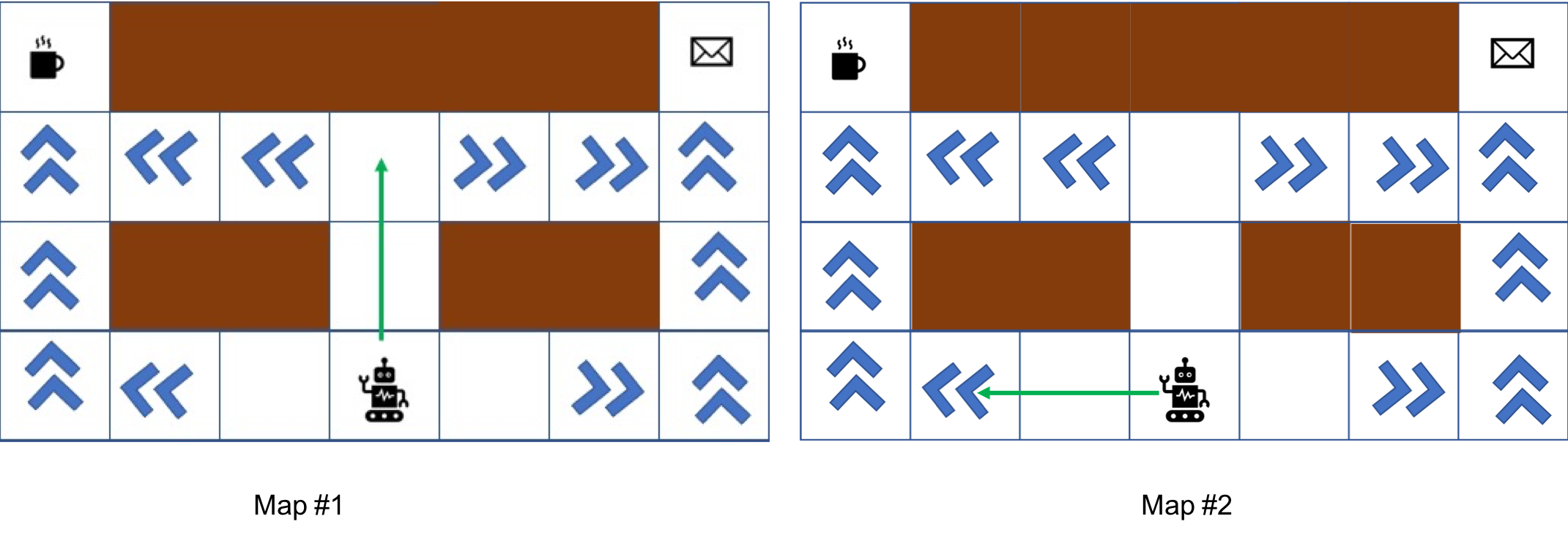}
\caption{Hypothesis 2.}
\label{study2}
\vspace{5pt}
\end{subfigure}
\begin{subfigure}[b]{\columnwidth}
\centering
\centering
\includegraphics[width=\columnwidth]{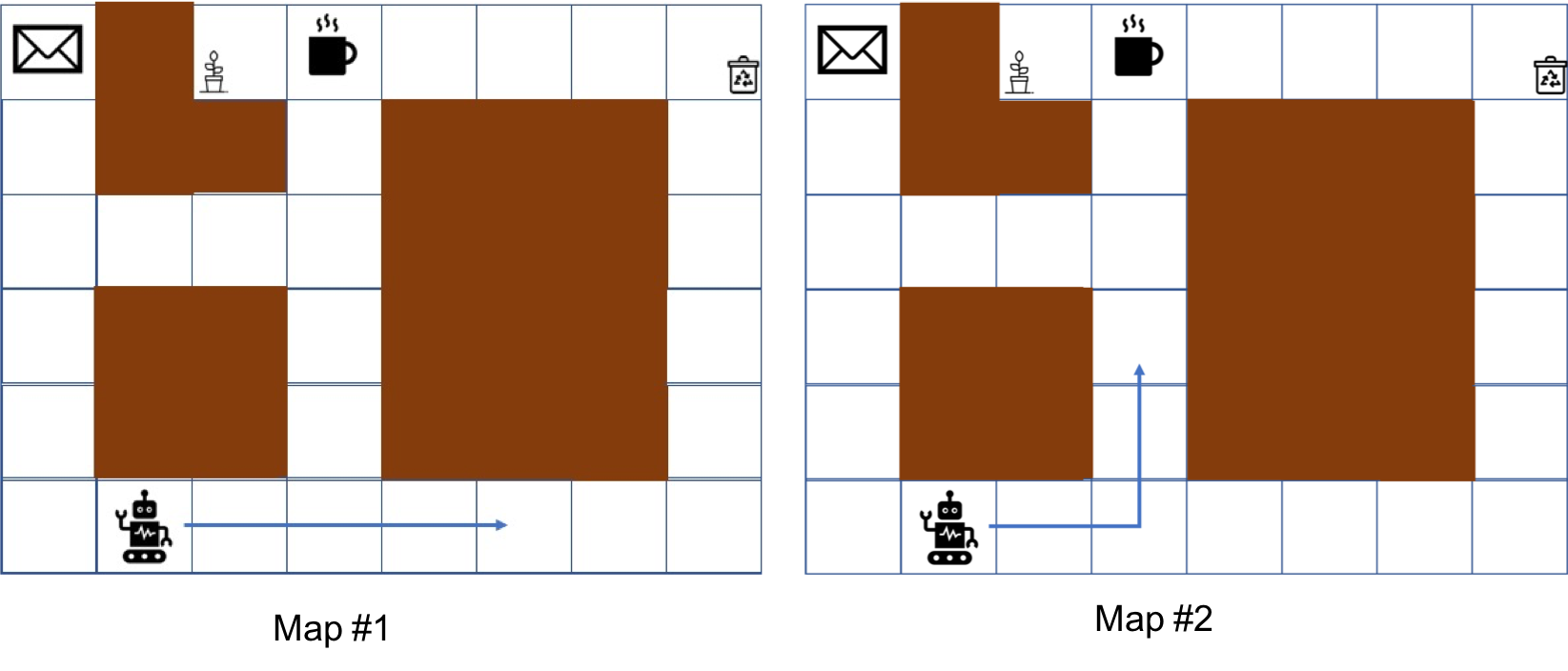}
\caption{Hypothesis 3.}
\label{study3}
\end{subfigure}
\caption{Images shown to users in the study.}
\label{fig:three-graphs}
\end{figure}

\subsection{Hypothesis 1}
\label{hypo1}

Our objective here is to ascertain how the presence or absence of 
other plans affects the explicability of a plan regardless of its 
distance from the expected plan. 
Our formulation asserts that the explicability of a plan will be 
lowered by plans that are closer to the expected plan in the human's model. 
To test this, we went with a within-subject study that showed the 
users two maps that are nearly identical except that 
one allows for more plans to the goal than the other (Figure \ref{study1}). 
The expected plan is the optimal plan. We also laterally shifted the position of the goal and the initial position of the robot. By our framework, the plan in Map \#1 should have lower explicability than the one in Map \#2.

\begin{figure}[!t]
\centering
\begin{subfigure}[b]{\columnwidth}
\centering
\includegraphics[width=\columnwidth]{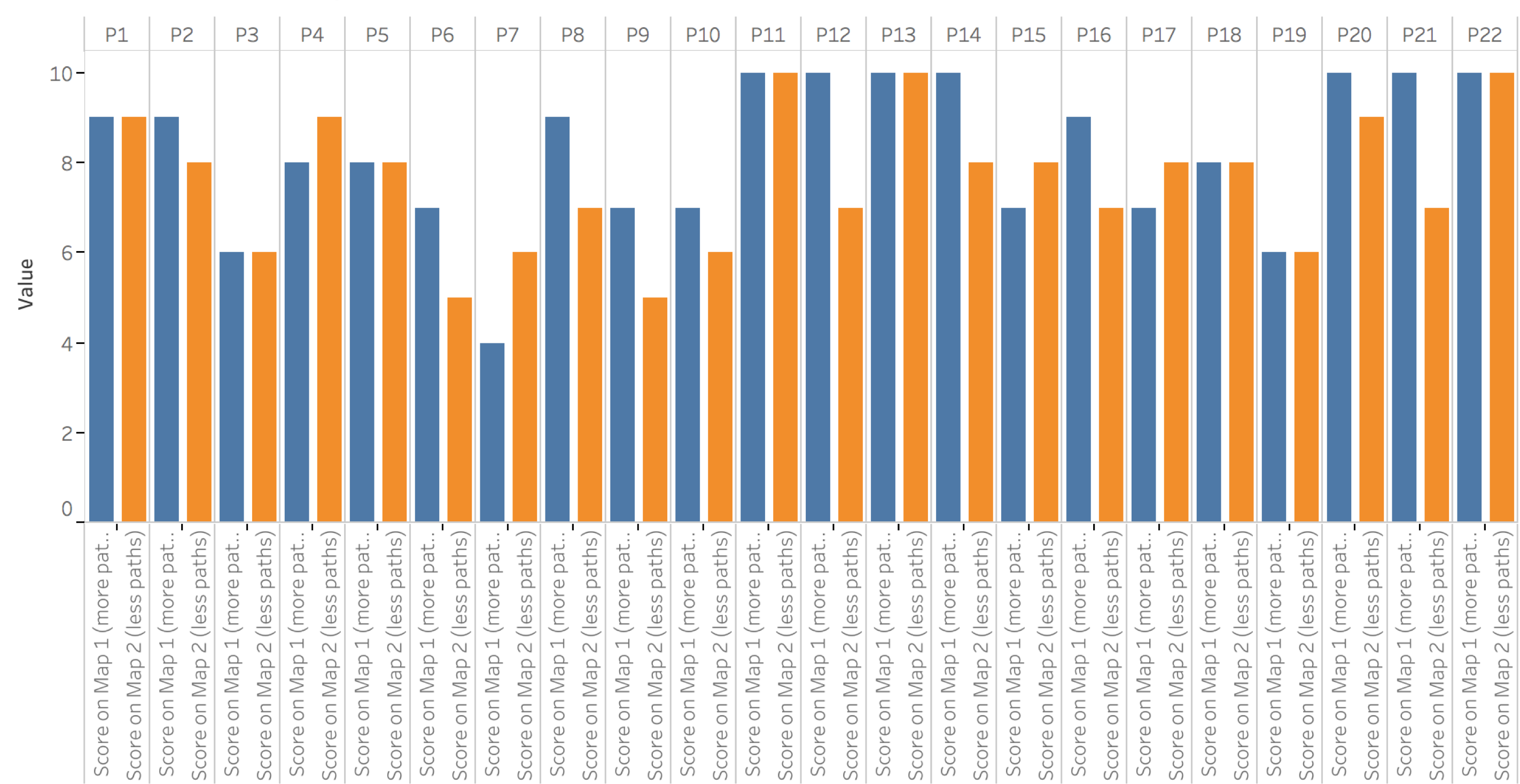}
\caption{Hypothesis 1.}
\label{study1-results}
\vspace{5pt}
\end{subfigure}
\begin{subfigure}[b]{\columnwidth}
\centering
\centering
\includegraphics[width=0.8\columnwidth]{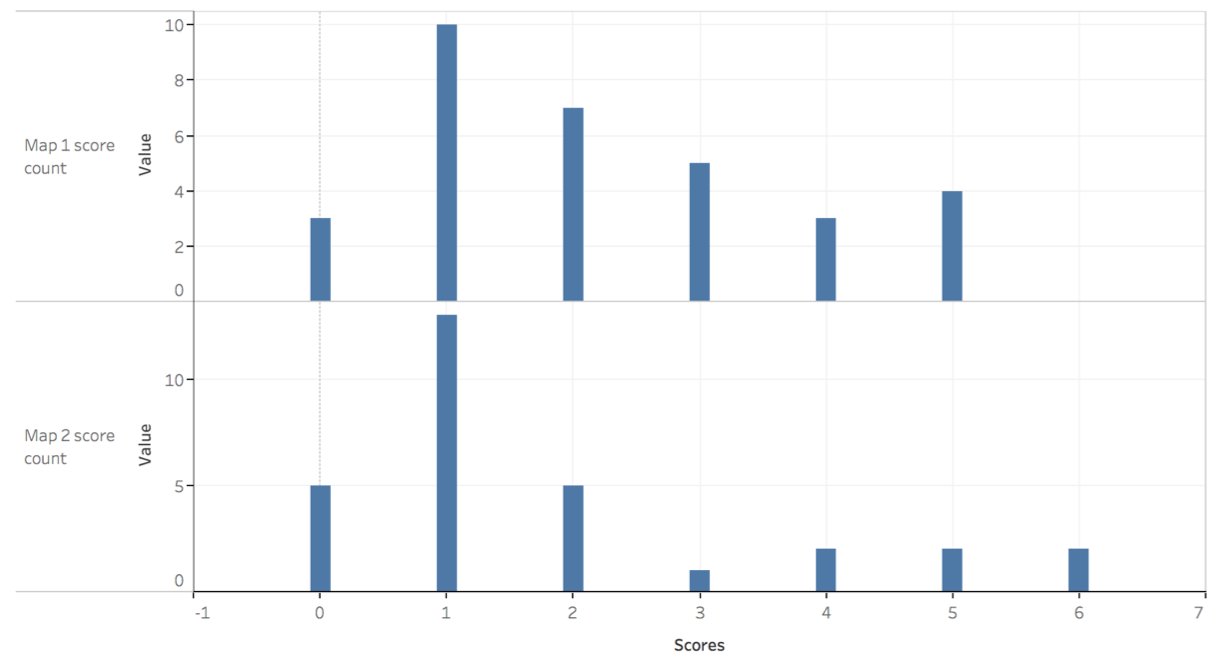}
\caption{Hypothesis 2.}
\label{study2-results}
\vspace{5pt}
\end{subfigure}
\begin{subfigure}[b]{\columnwidth}
\centering
\centering
\includegraphics[width=0.75\columnwidth]{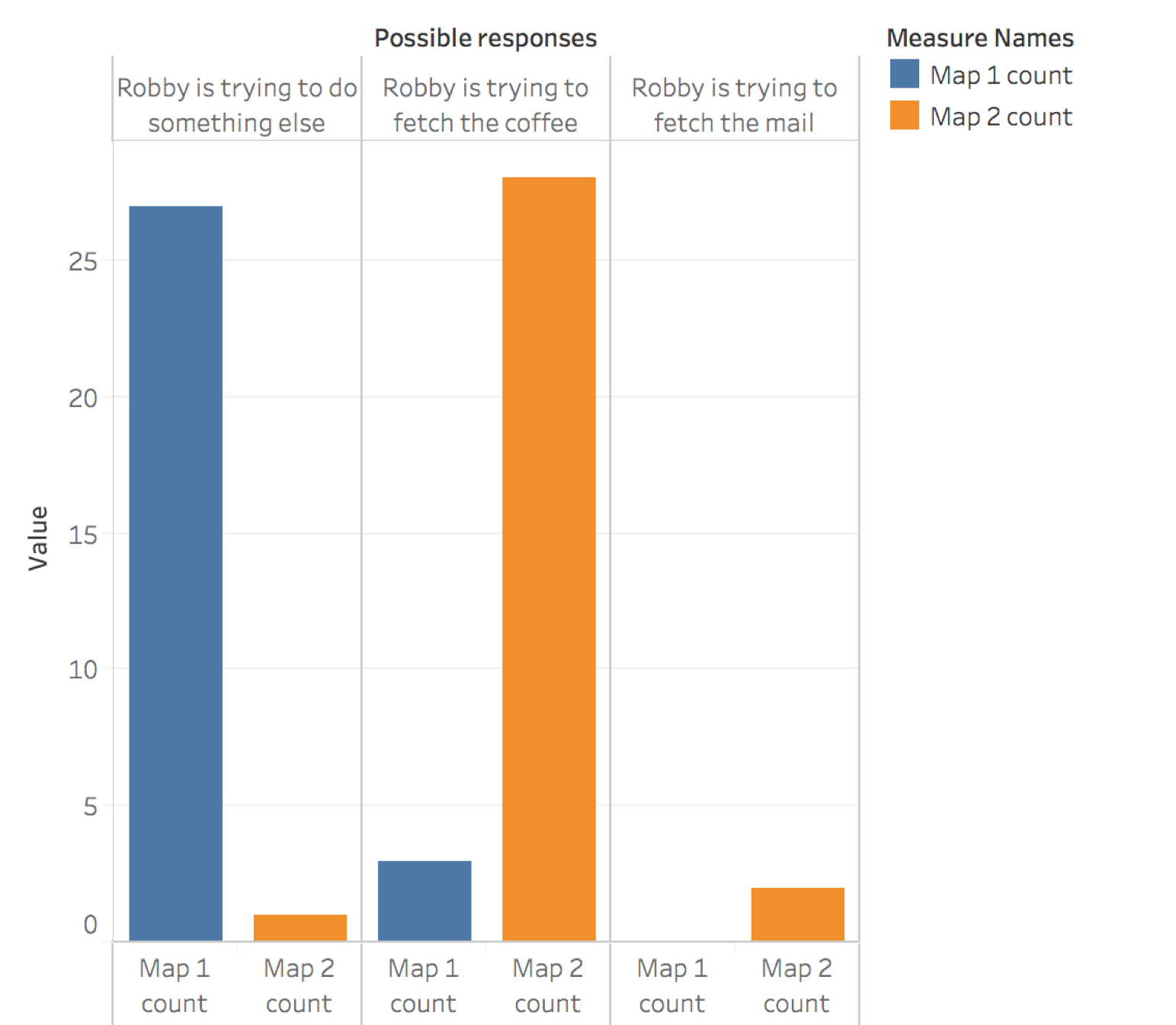}
\caption{Hypothesis 3.}
\label{study3-results}
\end{subfigure}
\caption{Filtered responses collected for each hypothesis.}
\label{fig:three-graphs-results}
\end{figure}

Each participant was shown the maps side-by-side and was asked to rate the plan on a scale from 0 to 10, where 0 was noted as being \textit{Completely Expected} and 10 being \textit{Completely Surprising}. 
As a filter question, we asked the user to describe their expected plan for each map and we filtered out any answers that did not refer to a straight line. 
This ensures that the results are with respect to people who would view both maps with respect to the similar expected plan. 
In this case, since the order of the maps could potentially affect the responses, 
we created two batches of the survey with different left-right ordering 
of the maps to ensure the results are counterbalanced. 
In total we collected 35 responses, one of which was ignored as the participant had taken part in a pilot study for the same map. Out of the remaining 34, we selected the 22 responses that had given the correct answer to the filter question. Within this set, the responses had an average score of 8.227 for the map with more plans (Map \#1 in Figure \ref{study1}) and a score of 7.591 for the other one. 
We ran a two tailed paired t-test and found a p-value of just 0.0401. This conforms to our hypothesis about the plan in Map \#1 being less explicable. Also note that we went with a paired t-test because of the within-subject study design where we had pairs of observation.


\subsection{Hypothesis 2}
\label{hypo2}

For the next hypothesis we validate the invariance of explicability to the 
behaviors that are consistent with one mental model versus multiple models. 
In this case, we went with a between-subject study (where each subject is only exposed to one condition) to avoid the possibility of subjects' beliefs about model likelihoods for one map being influenced by the other map. 
Here, each participant saw one of the maps listed in Figure \ref{study2}. 
They were told the robot is trying to either pick up the coffee, the mail, 
or that the robot may be trying to do something you do not know about. 
Then they were asked to mark their level of surprise about the plan. 
Since these are not paired scores, to avoid too much variance we went with a 
7-point Likert scale that again went from \textit{Completely Expected} 
to \textit{Completely Surprised}. 
To allow for more or less equal likelihood and thus similar explicability \footnote{With
additional constraints, there is a small difference 
in the probabilities associated with the prefixes in the two maps of Figure \ref{study2}
-- they come to 0.9912 and 0.9903 under priors 0.445, 0.445 and 0.01 
for models where the robot picks up coffee, mail and $\mathcal{M}^0$ 
respectively. The calculations are provided in Appendix \ref{calc}.}
among the plans, 
we added a few additional constraints to simplify the behavior space. 
In particular, we told the participants that the robot could not 
visit cells it has traveled to before, and for cells with direction arrows, 
the robot can only travel in that direction. With this constraint, the plan prefix in Map \#1 can lead to either of the goals, while in the case of Map \#2, the prefix only allows for the achievement of the goal of fetching coffee.
Here $\mathcal{M}^0$ was restricted to only consider behaviors that are 
feasible given the aforementioned constraints. 

We collected responses from 34 participants for Map \#1 and 35 participants 
for Map \#2. We asked them what goal they thought the robot was doing as a 
filter question. We considered only participants who answered either coffee 
and mail for Map \#1 and coffee and not mail for Map \#2, thereby ensuring that
the results accurately reflected the fact that the first set of responses are 
from people who believe that both models are possible and the second set are
from people who believe the prefix only corresponds to one model. We were left
with 32 and 31 entries for the two maps and had an average score of 2.218 for
Map \#1 and 1.867 Map \#2. After running t-test, we saw no strong
statistical evidence to believe that they come from significantly different
distributions (p-value of 0.4 which is generally regarded to be high probability for the null hypothesis). 
This conforms to our hypothesis.




\subsection{Hypothesis 3}
\label{hypo3}

Our third hypothesis is on the relationship between legibility and explicability, 
specifically that a prefix with low explicability should also have low legibility,
as entailed by our unified framework.
We again used a between-subject study design. Each subject was shown one plan prefix and made to guess the goal of the agent (Figure \ref{study3}). 
In this domain, to make the legibility calculation easier, in addition to the rules about the agent not being able to revisit previous cells, we added an additional rule that the robot can only move in three directions (forward, left, and right) and not backward. 
This makes sure that once the robot moves right far enough, it has zero probability of ever getting to mail and thus by traditional 
legibility methods should be taken as a strong evidence for the agent's goal being to pick up coffee. 

Of course, as per our hypothesis, in cases where the human in unsure about her knowledge this may no longer be the case. We induced this by (1) mentioning there may be other tasks that the robot may be interested in (2) adding visual distractions in the form of plant and a trash bin. Neither of these objects are referred to at any point in the instructions and were smaller in size than the other two objects (mail and coffee). Here mail and coffee were presented as possible goals for the robot. We believe this reflects a more realistic scenario as a robot would be expected to work alongside humans spaces which may be cluttered with everyday objects. 
Also to balance the placement of the objects, we place one unrelated object in the direction of each possible path.

For both maps, we collected 34 responses and used a filtering question
where participants were required to identify 
that the robot cannot revisit previously visited cells. 
For Map \#1, after filtering we were left with 30 responses (additionally, two responses were removed, as we had a reason to believe these submissions came from the same Turker; all the responses are provided in the appendix).
Out of 30 responses, 27 answered that they the 
objective of the robot was neither to pick up coffee nor mail, thereby showing the
path was not legible. For Map \#2 (which shows the explicable plan), after
filtering we were left with 31 responses, 28 out of which correctly identified 
the goal, which was to fetch coffee.
The results seem to align with our hypothesis that in these more general settings, one can't achieve legibility at the expense of explicability.
Thus the best approach to 
legibility may be to choose plans with relatively high explicability score that still help reveal the model.




\section{Conclusion and Discussion}
\label{conclusion}

Most works on interpretable behavior generation have focused on 
studying clean mathematical models that reflect our intuitions
about desirable behavioral properties. However, by focusing 
on individual behavioral properties we might overlook complexities that are essential for 
successful deployment of such methods in real-world applications. 
In this work, we introduced a framework that is able to explain 
deficiencies in existing methods. 
As we show through our experiments, our approach 
is able to correctly anticipate several properties of these methods
not studied in the prior literature. These properties only appear when the different aspects of interpretable behavior
are considered together, as one would when designing a 
human-aware agent. 
In the rest of this section, we will discuss a few more implications of 
the proposed formulation and directions for future work.

\subsubsection{Legibility and Explicability:}
Both these notions are related to the human's desire to recognize 
the model \cite{model-rec}). Our formulation shows that outside
limited cases, legibility and explicability cannot be fully isolated. 
Earlier works have been doing this by assuming away either legibility, like in explicability with the human's hypothesis consisting of a single model \cite{zhang2017plan,kulkarni2019explicable}, 
or by assuming away explicability by assigning zero prior on $\mathcal{M}^0$ for legibility \cite{dragan2013legibility,dragan2013generating,macnally2018action,kulkarni2019unified,SZ:MZaamas20}. 
Interestingly, in cases where the human 
is aware that the agent is trying to be legible, the human may be more open to suboptimal behavior from the agent as they might attribute it to trying to communicate.
However, this does not eliminate $\mathcal{M}^0$
but instead introduces a new level of nesting for reasoning: the human 
is trying to make sense of an agent that is expected to reason over their beliefs about the human. This comes with all the known complexities and pitfalls of reasoning with nested beliefs \cite{fagin2003reasoning}.

\subsubsection{Planning:} 
The next logical step for this work
would be to be able to generate plans with the metrics
in the unified framework. A good starting point may be to 
compile the problem to a classical planning problem,
as done in \cite{exact} for explicable plans.

\subsubsection{Longitudinal Interactions:}
\label{longitudinal}
Our formulation currently looks at interpretability metrics for 
one-off interactions only. 
In cases where a human interacts with the agent for a long period, 
we can expect the user to start with a uniform distribution over
hypotheses models and a low probability for $\mathcal{M}^0$. 
In order to take a more long-term view of the human's interaction 
with the same agent (say, over a time horizon), 
legibility and predictability measures can be handled by directly 
carrying over the posterior from each interaction to the next one. 
However, for explicability more care needs to be taken. 
For example, \citet{kulkarni2020designing} hypothesize 
a possible discounting of inexplicable behavior. 
They argue that after the first inexplicability, a human would
be less surprised when similar inexplicable behavior was again
presented to her. 
It would be interesting to see the implications of such longitudinal considerations on our model.

\subsubsection{Communication and Explanation:} 
Finally, each of the metrics can also be improved through 
explicit communication. For legibility and explicability, this can be 
done by providing model information, while for predictability, 
this can be done by providing information about the plan.  
Works like \citet{explain} can be viewed as
exclusively trying to use explanation as a process of providing information
that improves the explicability of a plan. 
However, in their formulation explanations are restricted to model information.
The reformulation of explicability in terms of a probabilistic framework gives
us the tools to include other kinds of explanatory information \cite{miller2019explanation,miller2018contrastive,xaip-landscape} 
in order to improve the likelihood: e.g.
by simplifying the model through the use of abstractions \cite{HELM,d3wa-exp,LIME,madumal2019explainable} or
by converting the problem into one of checking the likelihood of a related but simpler artifact like a causal chain \cite{seegebarth}
or an abstract policy \cite{topin2019generation}. 
The likelihood function can be chosen to also reflect the user's inherent computation limitations (e.g. noisy rational models). 

\section*{Acknowledgments}

The research of ASU contributors to the paper is supported in parts by ONR grants N00014-16-1-2892, N00014-18-1-2442, N00014-18-1-2840, N00014-9-1-2119, AFOSR grant FA9550-18-1-0067, DARPA SAIL-ON grant W911NF-19-2-0006, NASA grant NNX17AD06G, and a JP Morgan AI Faculty Research grant.

\bibliography{aaai_subm}

\clearpage
\section{Appendix}
\subsection{Probability Calculation For Hypothesis 2}
\label{calc}
In the setting described in Section \ref{hypo2}, the node $\mathbb{M}^R$ can take values from the set $\{\mathcal{M}^c,\mathcal{M}^m, \mathcal{M}^0\}$, where $\mathcal{M}^c$ is the model where the goal of the robot is to fetch coffee, $\mathcal{M}^m$ is the model where the goal is to fetch the mail and $\mathcal{M}^0$ is the random model. We assume a prior of $0.445$ for coffee and mail domains and prior of $0.01$ for $\mathcal{M}^0$. Given the constraints of the domain (i.e., the robot can't revisit cells and cells with direction symbols allow for robot's movement in only that direction), there are only two feasible paths in $\mathcal{M}^c$ and $\mathcal{M}^m$ of equal likelihood. While in the case of $\mathcal{M}^0$ it assigns a probability of nearly $0.0435$ to each feasible path (there are nearly 23 feasible paths in the domain). Also, we restrict $\hat{\tau}_{obs}$, to five step prefixes. Now for Map \#1 the probability calculations are as follows:
\begin{align}
    \begin{split}
  P(\mathcal{M}^c|\hat{\tau}_{Map\#1})+P(\mathcal{M}^m|\hat{\tau}_{Map\#1}) =\\ \frac{P(\hat{\tau}_{Map\#1}|\mathcal{M}^c)*P(\mathcal{M}^c) + P(\hat{\tau}_{Map\#1}|\mathcal{M}^m)*P(\mathcal{M}^m) }{P(\hat{\tau}_{Map\#1})}      
    \end{split}
\end{align}
Now we get $P(\hat{\tau}_{Map\#1}|\mathcal{M}^c) = P(\hat{\tau}_{Map\#1}|\mathcal{M}^m)= 0.5$, since in each case its one of the only two possible paths to be taken by the robot. To calculate $P(\hat{\tau}_{Map\#1})$,
\begin{align}
    \begin{split}
        P(\hat{\tau}_{Map\#1}) = P(\hat{\tau}_{Map\#1}|\mathcal{M}^c)*P(\mathcal{M}^c) +\\ P(\hat{\tau}_{Map\#1}|\mathcal{M}^m)*P(\mathcal{M}^m) + P(\hat{\tau}_{Map\#1}|\mathcal{M}^0)*P(\mathcal{M}^0)
    \end{split}
\end{align}
Since in $\mathcal{M}^0$ this could be one nine different completions, we assign $P(\hat{\tau}_{Map\#1}|\mathcal{M}^0) = 0.3914$. Thus $P(\hat{\tau}_{Map\#1}) = 0.448915$ and the total becomes $0.9912$.
For the path in Map \#2, we can repeat the calculations, the main difference here is that there is no completion for this prefix in $\mathcal{M}^m$ and thus it should have 0 probability in that model and in $\mathcal{M}^0$ there is only 5 completions and thus $P(\hat{\tau}_{Map\#1}|\mathcal{M}^0) = 0.2175$, thus the total probabilities come to $0.9903$.

\end{document}